\documentclass[letterpaper, 10 pt, conference]{ieeeconf}  % Comment this line out if you need a4paper

\IEEEoverridecommandlockouts                              % This command is only needed if 
\makeatletter
\let\NAT@parse\undefined
\makeatother

\usepackage{cite}
\usepackage{amsmath,amssymb,amsfonts}
\usepackage{algorithmic}
\usepackage{graphicx}
\usepackage{pifont}
\usepackage{textcomp}
\usepackage{hyperref}
\usepackage{multirow}
\usepackage{balance}
\usepackage[table,xcdraw]{xcolor}
\usepackage{url}
\usepackage{comment}
\usepackage{color}
\usepackage{latexsym}
\usepackage{mathtools}
\usepackage{caption}
\usepackage{subcaption}
\usepackage{orcidlink}

\DeclareMathOperator*{\argmax}{\arg\!\max} % argmax operator

\title{Mixed-Reality Digital Twins: Leveraging the Physical and Virtual Worlds for Hybrid Sim2Real Transition of Multi-Agent Reinforcement Learning Policies}

\author{Chinmay V. Samak$^{\ast}$ \orcidlink{0000-0002-6455-6716}, Tanmay V. Samak$^{\ast}$ \orcidlink{0000-0002-9717-0764} and Venkat N. Krovi\orcidlink{0000-0003-2539-896X}% <-this % stops a space
\thanks{$^{\ast}$These authors contributed equally.}% <-this % stops a space
\thanks{This work was supported in part by the U.S. National Science Foundation under NSF IIS-1925500 and NSF CNS-1939058.}% <-this % stops a space
\thanks{C. V. Samak, T. V. Samak, and V. N. Krovi are with the Department of Automotive Engineering, Clemson University International Center for Automotive Research (CU-ICAR), Greenville, SC 29607, USA. Email: {\tt\small {\{\href{mailto:csamak@clemson.edu}{csamak}, \href{mailto:tsamak@clemson.edu}{tsamak}, \href{mailto:vkrovi@clemson.edu}{vkrovi}\}@clemson.edu}}}% <-this % stops a space
}

\begin{document}

\maketitle
\thispagestyle{empty}
\pagestyle{empty}

%%%%%%%%%%%%%%%%%%%%%%%%%%%%%%%%%%%%%%%%%%%%%%%%%%%%%%%%%%%%%%%%%%%%%%%%%%%%%%%%

\begin{abstract}
Multi-agent reinforcement learning (MARL) for cyber-physical vehicle systems usually requires a significantly long training time due to their inherent complexity. Furthermore, deploying the trained policies in the real world demands a feature-rich environment along with multiple physical embodied agents, which may not be feasible due to monetary, physical, energy, or safety constraints. This work seeks to address these pain points by presenting a mixed-reality (MR) digital twin (DT) framework capable of: (i) boosting training speeds by selectively scaling parallelized simulation workloads on-demand, and (ii) immersing the MARL policies across hybrid simulation-to-reality (sim2real) experiments. The viability and performance of the proposed framework are highlighted through two representative use cases, which cover cooperative as well as competitive classes of MARL problems. We study the effect of: (i) agent and environment parallelization on training time, and (ii) systematic domain randomization on zero-shot sim2real transfer, across both case studies. Results indicate up to 76.3\% reduction in training time with the proposed parallelization scheme and sim2real gap as low as 2.9\% using the proposed deployment method.\\%
\end{abstract}

\begin{keywords}
Multi-Agent Systems, Autonomous Vehicles, Reinforcement Learning, Digital Twins, Real2Sim, Sim2Real\\%
\end{keywords}

%%%%%%%%%%%%%%%%%%%%%%%%%%%%%%%%%%%%%%%%%%%%%%%%%%%%%%%%%%%%%%%%%%%%%%%%%%%%%%%%

\section{Introduction}
\label{Section: Introduction}

Connected autonomous vehicles (CAVs) are exemplars of cyber-physical systems (CPS) operating within an environment with other agents. The development and deployment of such systems present a formidable challenge due to the increased complexity of multi-agent interactions. In such a milieu, multi-agent learning stands out as a promising avenue for developing autonomous vehicles capable of navigating complex and dynamic environments while considering the nature of interactions with their peers. Particularly, multi-agent reinforcement learning (MARL) offers the tantalizing potential of learning through self-exploration, which can potentially capture intricate cooperative/competitive multi-agent interactions.

\begin{figure}[t]
     \centering
     \begin{subfigure}[b]{\linewidth}
         \centering
         \includegraphics[width=\linewidth]{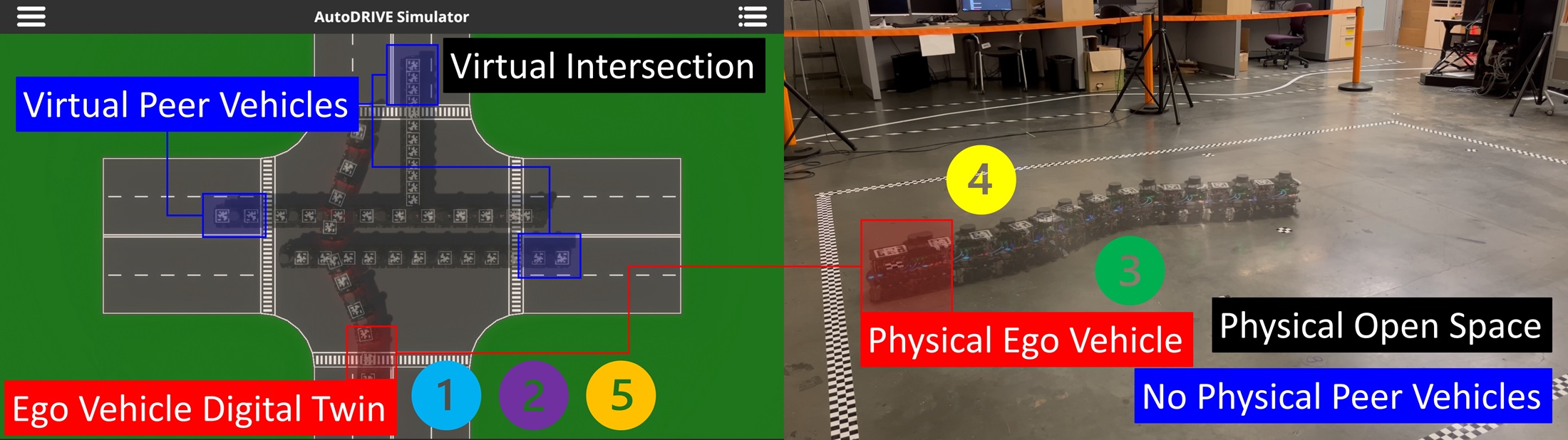}
         \caption{Cooperative MARL using Nigel.}
         \label{fig1a}
     \end{subfigure}
     \hfill
     \begin{subfigure}[b]{\linewidth}
         \centering
         \includegraphics[width=\linewidth]{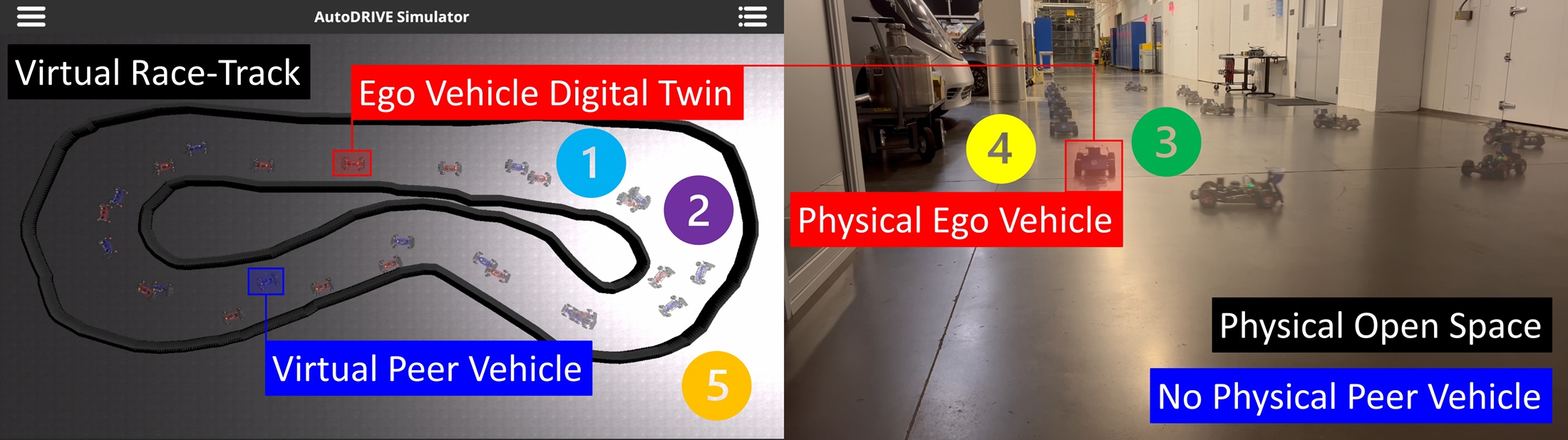}
         \caption{Competitive MARL using F1TENTH.}
         \label{fig1b}
     \end{subfigure}
     \caption{Proposed mixed-reality digital twin framework for hybrid sim2real transfer of MARL systems: (1) observe, (2) decide, (3) act, (4) estimate, and (5) update. \textbf{Video:} \small \url{https://youtu.be/ZHT34kwSe9U}}
    \label{fig1}
\end{figure}

\begin{figure*}[t]
     \centering
     \begin{subfigure}[b]{0.24\linewidth}
         \centering
         \includegraphics[width=\linewidth]{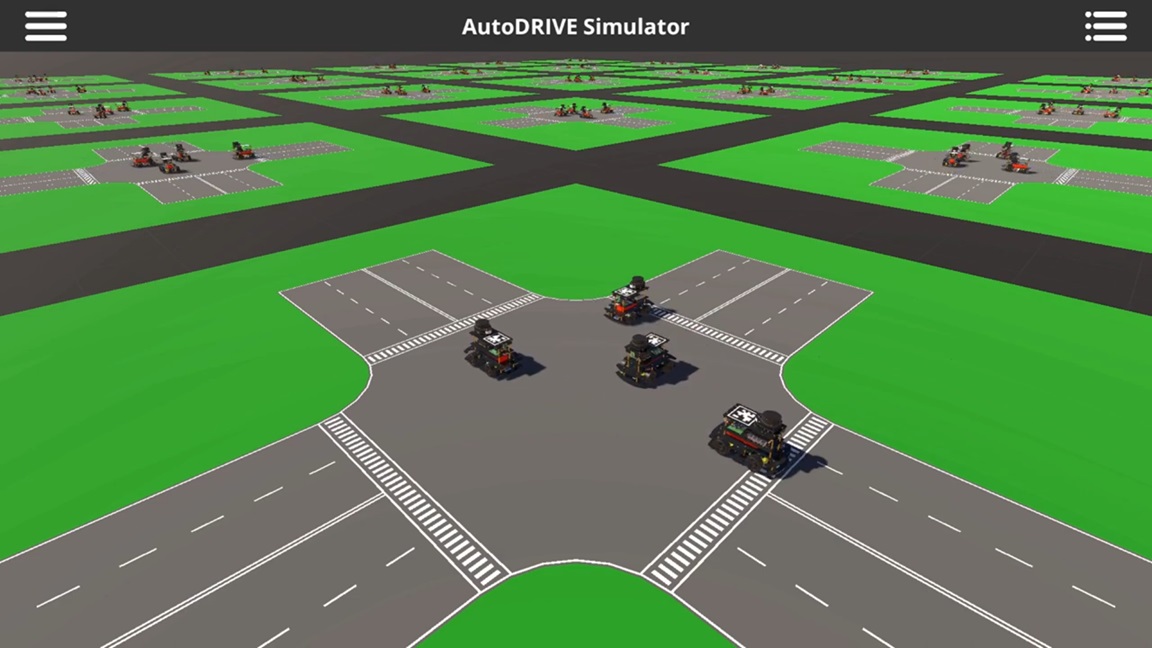}
         \caption{}
         \label{fig2a}
     \end{subfigure}
     \hfill
     \begin{subfigure}[b]{0.24\linewidth}
         \centering
         \includegraphics[width=\linewidth]{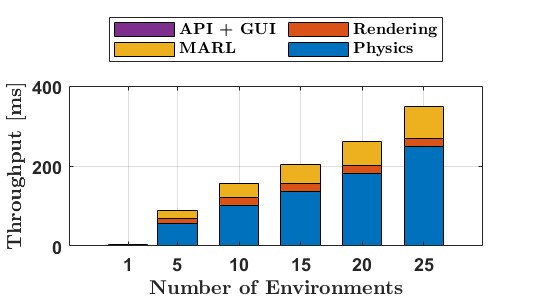}
         \caption{}
         \label{fig2b}
     \end{subfigure}
     \hfill
     \begin{subfigure}[b]{0.24\linewidth}
         \centering
         \includegraphics[width=\linewidth]{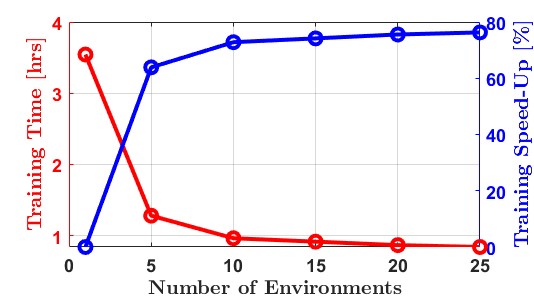}
         \caption{}
         \label{fig2c}
     \end{subfigure}
     \hfill
     \begin{subfigure}[b]{0.24\linewidth}
         \centering
         \includegraphics[width=\linewidth]{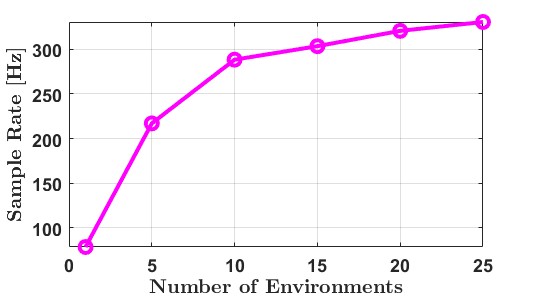}
         \caption{}
         \label{fig2d}
     \end{subfigure}
     \hfill
     \begin{subfigure}[b]{0.24\linewidth}
         \centering
         \includegraphics[width=\linewidth]{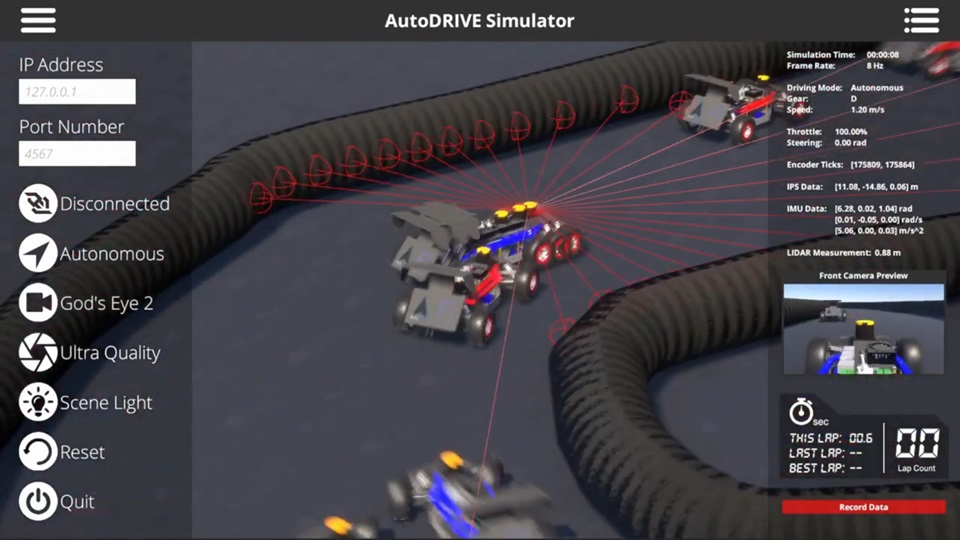}
         \caption{}
         \label{fig2e}
     \end{subfigure}
     \hfill
     \begin{subfigure}[b]{0.24\linewidth}
         \centering
         \includegraphics[width=\linewidth]{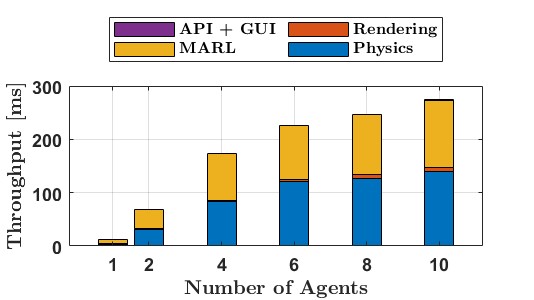}
         \caption{}
         \label{fig2f}
     \end{subfigure}
     \hfill
     \begin{subfigure}[b]{0.24\linewidth}
         \centering
         \includegraphics[width=\linewidth]{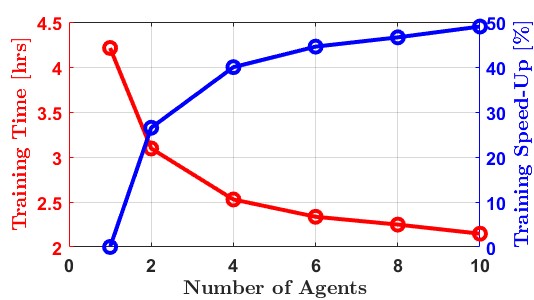}
         \caption{}
         \label{fig2g}
     \end{subfigure}
     \hfill
     \begin{subfigure}[b]{0.24\linewidth}
         \centering
         \includegraphics[width=\linewidth]{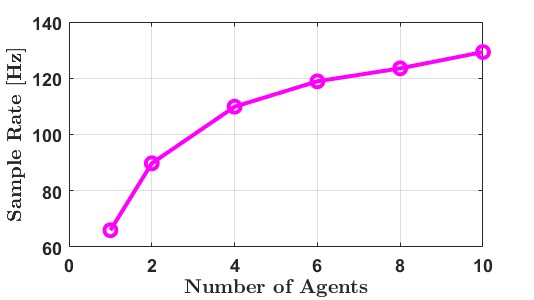}
         \caption{}
         \label{fig2h}
     \end{subfigure}
     \caption{Simulation parallelization: (a) 25$\times$4 cooperative MARL agents analyzed for (b) computational throughput, (c) training time, and (d) sample rate across different levels of environment parallelization; (e) 10$\times$2 competitive MARL agents analyzed for (f) computational throughput, (g) training time, and (h) sample rate across different levels of agent parallelization.}
    \label{fig2}
\end{figure*}

Cooperative MARL \cite{semnani2020multiagent, long2018optimally, 9316033, DuckietownMARL2022} fosters an environment where autonomous vehicles collaborate and share information to accomplish collective objectives such as optimizing traffic flow, enhancing safety, and efficiently navigating road networks. It mirrors traffic situations where vehicles must work together, such as intersection management or platooning scenarios. Challenges in cooperative MARL include coordinating vehicle actions to minimize congestion, maintaining safety margins, and ensuring smooth interactions with peers.

On the other hand, competitive MARL \cite{9790832, MultiTeamRacing2023, fuchs2020, song2021} introduces elements of privacy and rivalry in autonomous driving, simulating scenarios such as overtaking, merging in congested traffic, or racing. In this paradigm, agents strive to outperform their counterparts, prioritizing individual success over coordination. Challenges in competitive MARL encompass strategic decision-making, opponent modeling, and adapting to aggressive driving behaviors while preserving safety.

In either case, one of the key challenges in MARL training is longer wall-clock times, which can be accelerated \cite{DRLAcc2024, stooke2019acceleratedmethodsdeepreinforcement} by (a) developing sample-efficient RL algorithms \cite{pmlr-v178-kane22a}, or (b) accelerating simulations to improve data generation rate \cite{9494694}. This work focuses on the latter aspect, which in turn can be addressed by (i) training in low-fidelity simulations that can run faster than real-time \cite{semnani2020multiagent, long2018optimally, 9316033}, (ii) parallelizing simulations to accelerate data collection \cite{fuchs2020, song2021}, or (iii) both \cite{MultiTeamRacing2023}. Here, adopting low-fidelity simulations usually leads to a heightened sim2real gap, as marked in the literature through simulation-only deployments or explicit remarks for real-world experiments \cite{MultiTeamRacing2023}. Contrarily, adopting brute-force parallelization usually requires extensive computational resources \cite{fuchs2020, song2021, NVIDIAParallelRL2022}, which may not be sustainable -- this is because existing frameworks cannot \textit{``smartly''} parallelize replicas of multi-agent systems by selectively isolating collision, interaction, and perception between the agents/environments.

Another challenge arises when transitioning MARL policies from simulation to reality, either for training/fine-tuning or inference/evaluation. This is due to the requirement of multiple vehicles operating in a physical test environment, which may not always be feasible due to monetary, spatial, energy, or safety constraints. Consequently, there is a need to develop novel sim2real transfer techniques for MARL based on domain adaptation \cite{DomainAdaptation}, identification \cite{Identification1}, or augmentation \cite{DomainRandomization2017} methods, which have been explored in the context of single-agent RL. For example, \cite{9812407} proposes an interesting approach of using pre-recorded data to approximate ego viewpoint with a synthetic vehicle rendered as an obstacle during training. However, they use a simple kinematic model for agent simulation, and require a full-scale physical ado vehicle during testing, which can be prohibitively expensive, difficult to set up, and potentially unsafe. More recently, \cite{DuckietownMARL2022, MultiTeamRacing2023} have tried to address this gap, albeit partially, in the context of MARL. Particularly, \cite{DuckietownMARL2022} randomizes agent dynamics to mitigate the sim2real gap, but misses other potential domain randomization aspects like environment dynamics or agent observations/actions. Additionally, they do not consider simulation parallelization, and wait till the end of an episode to perform domain randomization. \cite{MultiTeamRacing2023} proposes system identification and domain randomization to mitigate the sim2real gap. They parallelize low-fidelity simulation instances for faster training, but without considering selective isolation of perception, collisions, and interactions of the agents, and consequently, they too have to wait till the end of an episode to perform domain randomization. While these approaches are stepping stones toward MARL sim2real transfer, they both simplified the problem by adopting extensive observation spaces with ground-truth information about the environment and employing mocap for sim2real transfer, which made the experimental setup complicated and expensive, with the trained policies heavily dependent on it. Additionally, both of them (typical of many others in literature) used multiple physical vehicles, albeit scaled, in a synthetically constructed physical test environment for real-world deployments, which may not scale well.

Our work tries to address these MARL pain points by proposing and offering a unified set of open-source\footnote{\textbf{GitHub:} \scriptsize \url{https://github.com/AutoDRIVE-Ecosystem/MRDT-MARL}} tools for multi-agent robotics problems that were previously scattered and underexplored in related fields. The key contributions of this work can be summarized as follows:
\begin{itemize}
    \item \textbf{Smart Parallelization:} We contribute a modular simulation parallelization framework, which allows selectively isolating the exteroceptive perception, collision, and interaction within or among MARL system(s).
    \item \textbf{Digital Twinning:} We introduce a bi-directional mixed-reality digital twinning framework to immerse a limited number of physical agents within a digital environment running virtual peer(s) to train/evaluate MARL policies.
    \item \textbf{Case Studies:} This work presents a cooperative non-zero-sum use-case of intersection traversal and a competitive zero-sum use-case of head-to-head autonomous racing. The agents are provided with realistically sparse observation spaces and employ onboard state estimation for real-time feedback during sim2real transfer.
    \item \textbf{MARL Analysis:} The proposed MARL case studies are benchmarked against a comparable set of baseline algorithms to assess their efficacy. We numerically evaluate the sim2real gap to analyze the effect of systematic domain randomization introduced in this work.
\end{itemize}

The remainder of this paper is organized as follows: Section \ref{Section: Case Studies} elucidates the two MARL case studies, including their mathematical formulation. Section \ref{Section: Methodology} describes the workflow adopted to train and deploy the MARL policies, including the mixed-reality digital twinning framework for sim2real transfer. Section \ref{Section: Results} analyzes the MARL training and deployment results for both case studies. Finally, Section \ref{Section: Conclusion} provides concluding remarks and future research directions.

%%%%%%%%%%%%%%%%%%%%%%%%%%%%%%%%%%%%%%%%%%%%%%%%%%%%%%%%%%%%%%%%%%%%%%%%%%%%%%%%%

\section{Case Studies}
\label{Section: Case Studies}

\subsection{Cooperative Multi-Agent Scenario}
\label{Sub-Section: Cooperative Multi-Agent Scenario}

We formulated a decentralized 4-agent collaborative scenario, wherein each agent's objective was to traverse a 2+2 lane, 4-way intersection without colliding or overstepping lane bounds. This scenario is representative of a standard uncontrolled traffic intersection. Here, each agent perceived its intrinsic states and received limited information from its peers (via V2V communication); no external sensing modalities were employed. Each agent was reset independently, resulting in highly stochastic initial conditions. The exact structure/map of the environment was not known to any agent. Consequently, this problem was framed as a partially observable Markov decision process (POMDP), which captured hidden state information through limited observations.

\subsubsection{Observation Space}
\label{Sub-Sub-Section: Observation Space I}

Each agent, $i$ ($0<i<N$), employed an appropriate subset of its sensor suite to collect observations: $o_{t}^{i} = \left [ g^{i}, \tilde{p}^{i}, \tilde{\psi}^{i}, \tilde{v}^{i} \right ]_{t} \in \mathbb{R}^{2+4(N-1)}$. This included IPS for positional coordinates $\left [ p_{x}, p_{y} \right ]_{t} \in \mathbb{R}^{2}$, IMU for yaw $\psi_{t} \in \mathbb{R}^{1}$, and incremental encoders for estimating vehicle velocity $v_{t} \in \mathbb{R}^{1}$. This follows that $g_{t}^{i} = \left [ g_{x}^{i}-p_{x}^{i}, g_{y}^{i}-p_{y}^{i} \right ]_{t} \in \mathbb{R}^{2}$ was the ego agent's goal location relative to itself, $\tilde{p}_{t}^{i} = \left [ p_{x}^{j}-p_{x}^{i}, p_{y}^{j}-p_{y}^{i} \right ]_{t} \in \mathbb{R}^{2(N-1)}$ was the position of every peer agent relative to the ego agent, $\tilde{\psi}_{t}^{i} = \psi_{t}^{j}-\psi_{t}^{i} \in \mathbb{R}^{N-1}$ was the yaw of every peer agent relative to the ego agent, and $\tilde{v}_{t}^{i} = v_{t}^{j} \in \mathbb{R}^{N-1}$ was the velocity of every peer agent. Here, $i$ represents the ego agent and $j \in \left [ 0, N-1 \right ]$ represents every other (peer) agent.

\subsubsection{Action Space}
\label{Sub-Sub-Section: Action Space I}

The Ackermann-steered vehicles were controlled using throttle and steering commands: $a_t^i = \left [ \tau_t^i, \delta_t^i \right ] \in \mathbb{R}^{2}$. From a collision avoidance perspective, the throttle command allowed agents to speed up/down, and the steering command allowed agents to dodge their peers.

\subsubsection{Reward Function}
\label{Sub-Sub-Section: Reward Function I}

Extrinsic reward was formulated as:
\begin{equation}
r_{t}^{i}  =  
\begin{cases}
r_{goal} & \text{if safe traversal} \\
-k_p * \left \| g_{t}^{i} \right \|_{2} & \text{if traffic violation} \\
k_r*(0.001+\left \| g_{t}^{i} \right \|_{2})^{-1} & \text{otherwise}
\end{cases}
\end{equation}

This function rewarded each agent with $r_{goal}=1$ for successfully traversing the intersection and penalized them proportional to their distance from the goal, represented as $k_p * \left \| g_{t}^{i} \right \|_{2}$, for collisions or lane-boundary violations. The agents were also continuously rewarded inversely proportional to their distance-to-goal, to negotiate the sparse-reward problem. The reward ($k_r=0.01$) and penalty ($k_p=0.425$) coefficients were tuned based on $g_{t}^{i}$ to ensure that the agents incurred a unit positive or negative reinforcement (i.e., $\approx \pm 1$) at the end of an average episode, to ensure stable training and avoid reward hacking.

\subsubsection{Optimization Problem}
\label{Sub-Sub-Section: Optimization Problem I}

The extrinsic reward function motivated each agent to maximize its expected future discounted reward by learning an optimal policy $\pi^*_\theta \left(a_t|o_t\right)$.
\begin{align}
\argmax_{\pi_\theta \left(a_t|o_t\right)} \quad &\mathbb{E}\left [ \sum_{t=0}^{\infty} \gamma^t r_t \right ]
\end{align}

\subsection{Competitive Multi-Agent Scenario}
\label{Sub-Section: Competitive Multi-Agent Scenario}

We formulated a 2-agent autonomous racing scenario, wherein each agent's objective was to minimize its lap time without colliding with the track or its opponent. This scenario is representative of a standard F1TENTH (a.k.a. RoboRacer) autonomous racing competition \cite{f1tenth-race}. Here, each agent collected its observations; no information was shared among the agents. The exact map of the environment was not known to any agent. Consequently, this problem was also framed as a POMDP. However, contrary to the cooperative MARL, this problem adopted a hybrid imitation-reinforcement learning architecture to guide the agents' exploration, thereby reducing training time. To this end, we recorded 5 laps worth of single-agent demonstrations for each agent by manually driving it in sub-optimal trajectories.

\subsubsection{Observation Space}
\label{Sub-Sub-Section: Observation Space II}

Each agent collected a vectorized observation: $o_t^i = \left [ v_t^i, m_t^i \right ] \in \mathbb{R}^{28}$. Here, $v_t^i \in \mathbb{R}^{1}$ represents the estimated forward velocity of $i$-th agent, and $m_t^i = \left [ ^{1}m_t^i, ^{2}m_t^i, \cdots, ^{27}m_t^i \right ] \in \mathbb{R}^{27}$ is the LIDAR range array with 27 measurements uniformly spaced across 270$^{\circ}$ and split around each side of the heading vector, within a 10 m radius.

\subsubsection{Action Space}
\label{Sub-Sub-Section: Action Space II}

The action space to control Ackermann-steered vehicles was $a_t^i = \left [ \tau_t^i, \delta_t^i \right ] \in \mathbb{R}^{2}$. Here, the throttle command allowed the agents to optimize their speed profile, and the steering command allowed them to optimize their race line, overtake their peers, and avoid collisions.

\subsubsection{Reward Function}
\label{Sub-Sub-Section: Reward Function II}

Following signals guided the agents:

\begin{itemize}
	\item \textbf{Behavioral Cloning:} The behavioral cloning (BC) \cite{bain1995} algorithm updated the policy in a supervised fashion with respect to the recorded demonstrations, mutually exclusive of the reinforcement learning update.
	\item \textbf{GAIL Reward:} The generative adversarial imitation learning (GAIL) reward \cite{ho2016} $^{g}r_t$ ensured that the agent optimized its actions safely (mimicking safe demos) and ethically (avoiding proactive reward hacking) by promoting the closeness of new observation-action pairs to those from the recorded demonstrations.
	\item \textbf{Curiosity Reward:} The curiosity reward \cite{pathak2017} $^{c}r_t$ promoted exploration by rewarding proportional to the difference in predicted and actual encoded observations.
	\item \textbf{Extrinsic Reward:} The objective of lap time reduction and incorporation of motion constraints was handled using an extrinsic reward function $^{e}r_t$. The agents received a reward of $r_{checkpoint}=0.01$ for passing each of the 19 checkpoints $c_i$ on the race track, $r_{lap}=0.1$ upon completing a lap, $r_{best\:lap}=0.7$ upon achieving a new best lap time, and a penalty of $r_{collision}=-1$ for colliding with the track bounds or peer agent (in which case both agents were penalized equally). Additionally, a continuous reward promoted higher velocities $v_t$.
    \begin{equation}
    ^{e}r_t^i  =  
    \begin{cases}
    r_{collision} & \text{if collision} \\
    r_{checkpoint} & \text{if checkpoint passed} \\
    r_{lap} & \text{if lap completed} \\
    r_{best\:lap} & \text{if best lap time} \\
    0.01*v_t^i & \text{otherwise}
    \end{cases}
    \end{equation}
\end{itemize}

\subsubsection{Optimization Problem}
\label{Sub-Sub-Section: Optimization Problem II}

The multi-objective problem of maximizing the expected future discounted reward while minimizing the behavioral cloning loss $\mathcal{L}_{BC}$ is defined as:
\begin{align}
\argmax_{\pi^i_\theta \left(a_t|o_t\right)} \quad & \eta \left(\mathbb{E}\left [ \sum_{t=0}^{\infty} \gamma^t r^i_t \right ] \right) - (1-\eta) \mathcal{L}_{BC}
\end{align}
where $\eta$ weighs the degree of imitation and reinforcement learning updates, and $r^i_t =\, ^{g}r^i_t + ^{c}r^i_t + ^{e}r^i_t$.

%%%%%%%%%%%%%%%%%%%%%%%%%%%%%%%%%%%%%%%%%%%%%%%%%%%%%%%%%%%%%%%%%%%%%%%%%%%%%%%%%

\section{Methodology}
\label{Section: Methodology}

In this work, we adopted and adapted the AutoDRIVE Ecosystem \cite{AutoDRIVE2023} to model, simulate, train, and deploy two MARL case studies. This choice was driven based on the comparative analysis presented in \cite{AutoDRIVE2023}, which satisfied all the requirements of this study. From a digital twinning perspective, data-driven system identification and calibration were used to customize models of Nigel \cite{Nigel2024} and F1TENTH \cite{F1TENTH2019} vehicles from real-world data to ensure reliable simulation.

\subsection{Simulation Parallelization}
\label{Sub-Section: Simulation Parallelization}

We leveraged the open-source nature of AutoDRIVE Simulator to implement a selectively scalable agent/environment parallelization framework. The simulator was configured to take advantage of CPU multi-threading as well as GPU instancing (only if available) to efficiently parallelize various simulation objects and processes while maintaining cross-platform support. Following is an overview of the simulation parallelization schemes:

\begin{itemize}
    \item \textbf{Parallel Instances:} Multiple instances of the simulator application can be spun up to train families of multi-agent systems, each isolated within its own simulation instance. This is a brute-force parallelization technique, which can cause unnecessary computational overhead.
    \item \textbf{Parallel Environments:} Isolated agents can learn the same task in parallel environments, within the same simulation instance (refer Fig. \ref{fig2a}). This method can help train single/multiple agents in different environmental conditions, with slight variations in each environment.
    \item \textbf{Parallel Agents:} Parallel agents can learn the same task in the same environment, within the same simulation instance (refer Fig. \ref{fig2e}). The parallel agents may collide/perceive/interact with selective peers/opponents. Additionally, the parallel agents may or may not be exactly identical, thereby robustifying them against minor parametric variations.
\end{itemize}

It should be noted that parallelization beyond a certain point can hurt (i.e., point of diminishing return), wherein the increased simulation workload may slow down the training so much that parallel policy optimization can no longer accelerate it. This \textit{``saturation point''} is dependent on the hardware/software configuration, and is subject to change.

\subsection{Learning Architecture}
\label{Sub-Section: Learning Architecture}

While there is no limitation on RL algorithms to be employed within our framework, we leverage proximal policy optimization (PPO) \cite{PPO2017} for our MARL case-studies. PPO is an on-policy method, which is empirically equally effective as its off-policy counterparts \cite{MAPPO2022}. Moreover, PPO promotes stable and efficient learning by imposing 2 complementary constraints: (a) a clipped surrogate objective to control each action probability update, and (b) a KL divergence early stopping criterion to limit overall policy change.

In terms of policy updates, cooperative MARL uses the collective experience of all agents to update a common policy, a.k.a. centralized training and decentralized execution (CTDE) or multi-agent PPO (MAPPO). Contrarily, competitive MARL uses the independent experience of each agent to update its own policy, a.k.a. decentralized learning or independent PPO (IPPO), since the agents are in pure competition and not allowed to share any observations or experiences. Nevertheless, in both cases, the parallelized agents contribute their experiences to update their respective herd's policy. This results in distributed sampling, which improves data collection speed and diversity (refer Fig. \ref{fig2d} and Fig. \ref{fig2h}), thereby increasing their correlation with the true state-action distribution, and stabilizing the training.

Table \ref{tab1} hosts the detailed training configurations adopted for the cooperative as well as competitive MARL scenarios. The noted parameter values were arrived at by analyzing the agent(s)' behaviors to satisfy the intended objectives qualitatively, while also ensuring a stable learning process. MARL training was carried out on a single laptop PC with 12th Gen Intel Core i9-12900H 2.50 GHz CPU, NVIDIA GeForce RTX 3080 Ti GPU, and 32.0 GB RAM.

\begin{table}[t]
	\caption{Training Configurations}
	\begin{center}
        \resizebox{\columnwidth}{!}{%
		\begin{tabular}{l|l|l}
			\hline
			\multicolumn{1}{c|}{\textbf{\begin{tabular}[c]{@{}c@{}}PARAMETER\\ DESCRIPTION\end{tabular}}} &
                \multicolumn{1}{c|}{\textbf{\begin{tabular}[c]{@{}c@{}}COOPERATIVE\\ MARL\end{tabular}}} &
                \multicolumn{1}{c}{\textbf{\begin{tabular}[c]{@{}c@{}}COMPETITIVE\\ MARL\end{tabular}}} \\ \hline
			\multicolumn{2}{l}{\textbf{Hyperparameters}}                      \\ \hline
			Neural network architecture*               & \multicolumn{2}{l}{3-layer FCNN $\times$ \{128, Swish \cite{ramachandran2017}\}} \\
			Batch size                                & 64           & 64            \\
			Buffer size                               & 1024         & 1024          \\
			Learning rate ($\alpha$)                  & 3e-4         & 3e-4          \\
			Learning rate schedule                    & Linear       & Linear        \\
			Entropy regularization ($\beta$)          & 1e-3         & 1e-3          \\
			Policy update ($\epsilon$)                & 2e-1         & 2e-1          \\
			Regularization parameter ($\lambda$)      & 9.8e-1       & 9.8e-1        \\
			Epochs                                    & 3            & 3             \\
			Maximum steps ($n_{max}$)                 & 1e6          & 1e6           \\ \hline
			\multicolumn{2}{l}{\textbf{Behavioral Cloning}}                          \\ \hline
			Strength ($\eta$)                         & --           & 5e-1          \\ \hline
			\multicolumn{2}{l}{\textbf{GAIL Reward}}                                 \\ \hline
			Discount factor ($^{g}\gamma$)            & --           & 9.9e-1        \\
			Strength                                  & --           & 1e-2          \\
			Encoding size                             & --           & 128           \\
			Learning rate ($^{g}\alpha$)              & --           & 3e-4          \\ \hline
			\multicolumn{2}{l}{\textbf{Curiosity Reward}}                            \\ \hline
			Discount factor ($^{c}\gamma$)            & --           & 9.9e-1        \\
			Strength                                  & --           & 2e-2          \\
			Encoding size                             & --           & 256           \\
			Learning rate ($^{c}\alpha$)              & --           & 3e-4          \\ \hline
			\multicolumn{2}{l}{\textbf{Extrinsic Reward}}                            \\ \hline
			Discount factor ($^{e}\gamma$)            & 9.9e-1       & 9.9e-1        \\
			Strength                                  & 1.0          & 1.0           \\ \hline
            \multicolumn{3}{l}{*There are 2 identical networks: one acts as actor (policy) and the other as critic (value function).}
		\end{tabular}
        }
	\end{center}
	\label{tab1}
\end{table}

\subsection{Domain Randomization}
\label{Sub-Section: Domain Randomization}

We leveraged the simulation parallelization architecture to introduce systematic domain randomization \cite{DomainRandomization2017} across $k$ agent/environment replicas. Particularly, each parallelized agent/environment was simulated with a different set of dynamical parameters during each episode (more diverse domain randomization in less time), and since the parallelized agents/environments were mutually isolated from each other, this did not interrupt numerical integration between simulation time steps (i.e., maintained solver consistency). This is a slightly different strategy compared to those explored in the literature: (i) per-interval domain randomization (e.g., \cite{makoviychuk2021isaac}), which may violate solver consistency, or (ii) per-episode domain randomization (e.g., \cite{DuckietownMARL2022, MultiTeamRacing2023, 9562079}), which may limit diversity of parameters, even if parallelized.

Table \ref{tab2} hosts the detailed domain randomization parameters for the cooperative as well as competitive MARL scenarios. Particularly, since the cooperative MARL scenario was environment-parallelized, we vary the dynamics of each environment replica in this case. Contrarily, since the competitive MARL scenario was agent-parallelized, we vary the dynamics of each agent replica in this case. Additionally, in both cases, we also introduce noise in the agents' observations and actions at each time step. Here, the parameter $\xi$ denotes the degree of domain randomization. In this work, we analyze the effect of no domain randomization (NDR), i.e. $\xi = 0$, low domain randomization (LDR), i.e. $\xi = 1$, and high domain randomization (HDR), i.e. $\xi = 2$.

\begin{table}[t]
	\caption{Domain Randomization}
	\begin{center}
        \resizebox{\columnwidth}{!}{%
		\begin{tabular}{l|l|l}
			\hline
			\multicolumn{1}{c|}{\textbf{\begin{tabular}[c]{@{}c@{}}PARAMETER\\ DESCRIPTION\end{tabular}}} &
                \multicolumn{1}{c|}{\textbf{\begin{tabular}[c]{@{}c@{}}COOPERATIVE\\ MARL\end{tabular}}} &
                \multicolumn{1}{c}{\textbf{\begin{tabular}[c]{@{}c@{}}COMPETITIVE\\ MARL\end{tabular}}}                                                     \\ \hline
			\multicolumn{2}{l}{\textbf{Observation Noise}}                                                                                                  \\ \hline
			Position ($w_x^k$, $w_y^k$)                                      & $\xi \cdot N$(0,1e-4) m            & --                                      \\
			Orientation ($w_\psi^k$)                                         & $\xi \cdot N$(0,3.0625e-4) rad     & --                                      \\
			Velocity ($w_v^k$)                                               & $\xi \cdot N$(0,1e-4) m/s          & $\xi \cdot N$(0,1e-4) m/s              \\
			LIDAR Scan ($w_m^k$)                                             & --                                 & $\xi \cdot N$(0,1e-6) m                \\ \hline
			\multicolumn{2}{l}{\textbf{Action Noise}}                                                                                                       \\ \hline
			Throttle ($w_\tau^k$)                                            & $\xi \cdot N$(0,2.5e-3) norm\%     & $\xi \cdot N$(0,2.5e-3) norm\%         \\
            Steering ($w_\delta^k$)                                          & $\xi \cdot N$(0,2.5e-3) norm\%     & $\xi \cdot N$(0,2.5e-3) norm\%         \\ \hline
			\multicolumn{2}{l}{\textbf{Agent Dynamics}}                                                                                                     \\ \hline
			Center of Mass ($w_{x_{cg}}^k$, $w_{y_{cg}}^k$, $w_{z_{cg}}^k$)  & --                                 & $\xi \cdot$[-5e-2:1.11e-2:5e-2] m       \\
			Suspension Stiffness ($w_K^k$)                                   & --                                 & $\xi \cdot$[-100:22.22:100] N/m         \\
            Tire Stiffness ($w_{c_\alpha}^k$)                                & --                                 & $\xi \cdot$[-2.5:5.6e-1:2.5] N/rad      \\ \hline
			\multicolumn{2}{l}{\textbf{Environment Dynamics}}                                                                                               \\ \hline
			Surface Friction ($w_\mu^k$)                                     & $\xi \cdot$[-1e-1:8.33e-3:1e-1]    & --                                      \\
			Communication Delay ($w_d^k$)                                    & $\xi \cdot$[0:4.17e-4:1e-2] s      & --                                      \\ \hline
		\end{tabular}
        }
	\end{center}
	\label{tab2}
\end{table}

\subsection{Mixed-Reality Digital Twinning}
\label{Sub-Section: Mixed-Reality Digital Twinning}

\begin{figure}[t]
\includegraphics[width=\linewidth]{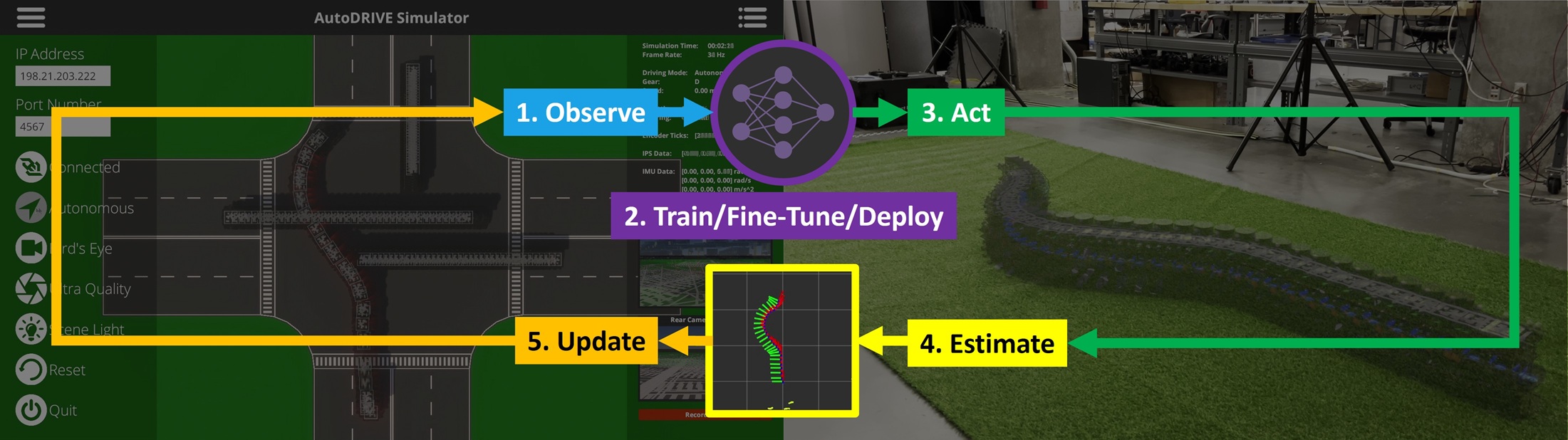}
\caption{Immersing MARL policies in the mixed-reality digital twin framework for training, fine-tuning, or deployment.}
\label{fig3}
\end{figure}

We propose a hybrid method for transferring the MARL policies from simulation to reality. The term \textit{``hybrid''} specifically alludes to the mixed-reality digital twin framework, which establishes a real-time bi-directional synchronization between the physical and virtual worlds. The intention is to minimize the number of physical agent(s) and environmental element(s) while deploying and validating MARL systems in the real world. Fig. \ref{fig1} (captured at 1 Hz) depicts the sim2real transfer of the trained MARL policies discussed in this work using the proposed framework while Fig. \ref{fig3} (captured at 5 Hz) depicts the possibility of optionally training/fine-tuning MARL policies (e.g., if there is a significant modification in the real-world setup such as the deliberately introduced turf mat in our case) within the same framework (thereby minimizing the experimental setup while enjoying the benefits of real-world data for policy update). Here, we deploy a single physical agent in an open space and connect it with its digital twin. The ``ego'' digital twin operates in a virtual environment with virtual peers, collects observations, optimizes (optionally, during training/fine-tuning) and/or uses (during testing/inference) the MARL policy to plan actions in the digital space. The planned action sequences are relayed back to the physical twin to be executed in the real world, which updates its state in reality. Finally, the ego digital twin is updated based on real-time state estimates of its physical twin (estimated on board) to close the loop. This process is repeated recursively until the experiment is completed. This way, we can exploit the real-world characteristics of vehicle dynamics and tire-road interactions while being resource-altruistic by augmenting environmental element(s) and peer agent(s) in the digital space. This also alleviates the safety concern of the experimental vehicles colliding with each other or the environmental element(s), especially as operational scales and number of agents increase.

%%%%%%%%%%%%%%%%%%%%%%%%%%%%%%%%%%%%%%%%%%%%%%%%%%%%%%%%%%%%%%%%%%%%%%%%%%%%%%%%%

\section{Results}
\label{Section: Results}

We benchmark MARL policies discussed in this work against 3 baselines. First, we choose follow-the-gap method (FGM) \cite{FGM2012} as a common benchmark for both cooperative and competitive tasks. Additionally, we benchmark the cooperative MARL policies against artificial potential field (APF) method \cite{APF1985} and timed-elastic-band (TEB) planner \cite{TEB2017}, which are common approaches for dynamic obstacle avoidance. Similarly, we also benchmark the competitive MARL policies against disparity-extender algorithm (DEA) \cite{DEA2019} and pure behavioral cloning (PBC) \cite{bain1995}, which are popular approaches in F1TENTH autonomous races. Finally, we also benchmark the performance of the best cooperative MARL policy before (base) and after fine-tuning (FT) in the real world to adapt to the deliberately introduced turf mat.

\subsection{Cooperative Multi-Agent Scenario}
\label{Sub-Section: Cooperative MARL Results}

\subsubsection{Training and Simulation Parallelization}
\label{Sub-Sub-Section: Training I}

\begin{figure}[t]
     \centering
     \begin{subfigure}[b]{0.32\linewidth}
         \centering
         \includegraphics[width=\linewidth]{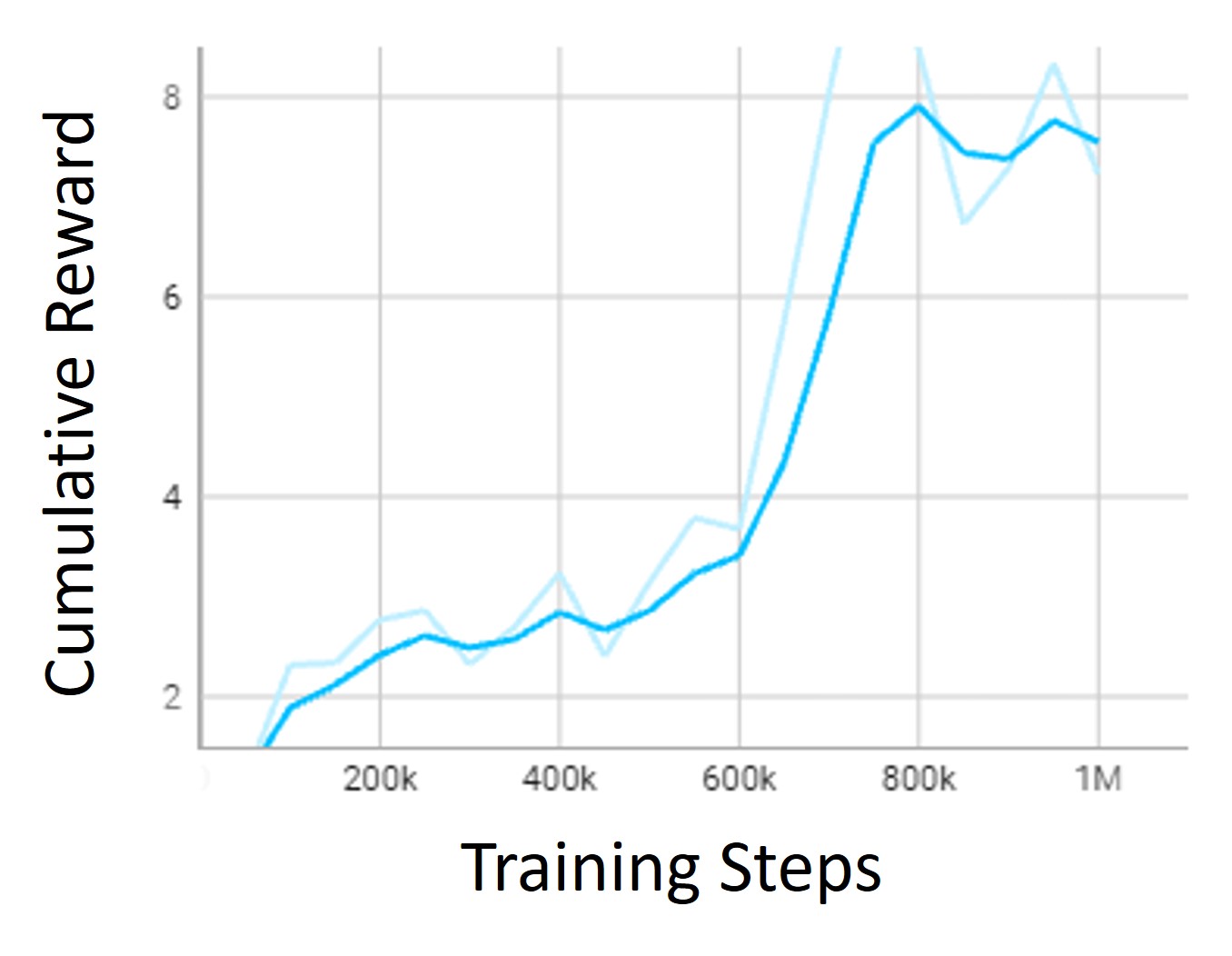}
         \caption{}
         \label{fig4a}
     \end{subfigure}
     \hfill
     \begin{subfigure}[b]{0.32\linewidth}
         \centering
         \includegraphics[width=\linewidth]{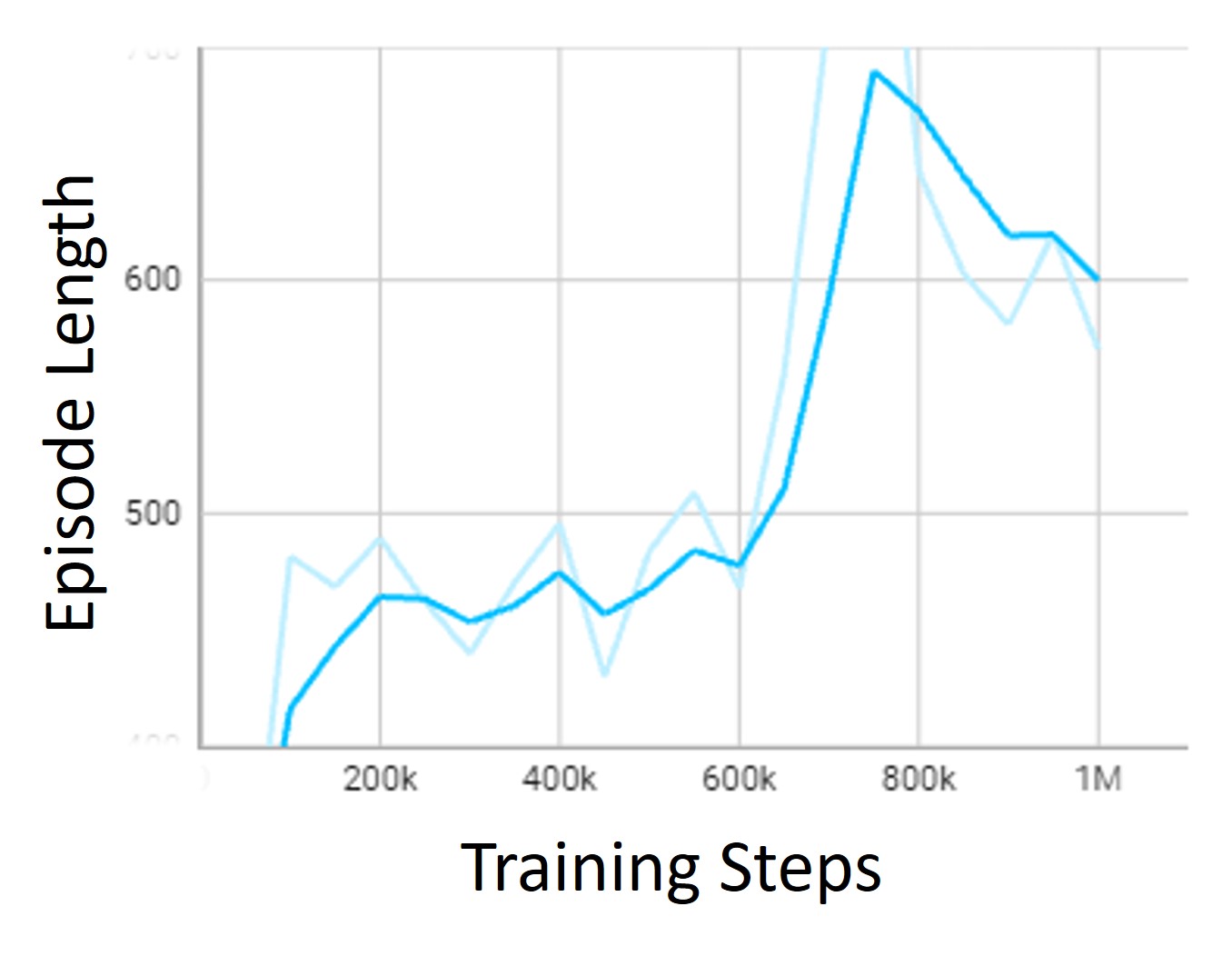}
         \caption{}
         \label{fig4b}
     \end{subfigure}
     \hfill
     \begin{subfigure}[b]{0.32\linewidth}
         \centering
         \includegraphics[width=\linewidth]{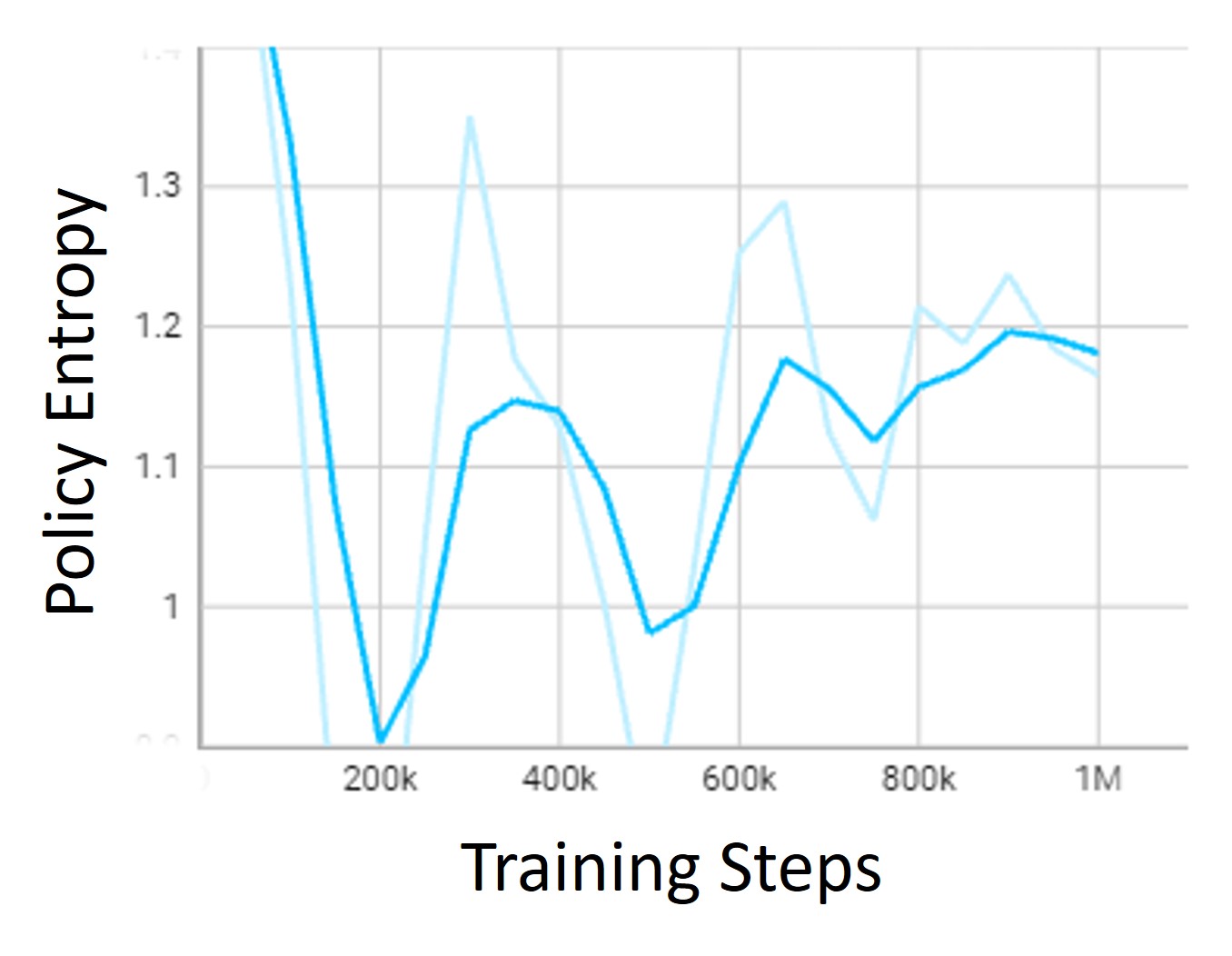}
         \caption{}
         \label{fig4c}
     \end{subfigure}
     \caption{Cooperative MARL training: (a) cumulative reward, (b) episode length, and (c) policy entropy w.r.t. training steps.}
    \label{fig4}
\end{figure}

Fig. \ref{fig4} depicts the key performance indicators (KPIs) used to analyze the cooperative MARL training without any domain randomization (i.e., NDR). It was observed that the agents took over 600k steps to understand the collective objective of safe intersection traversal. This is marked by a sustainable increase in the cumulative reward (from $\sim$3 to $\sim$8) as well as episode length (from $\sim$470 to $\sim$600 steps). This is also when the policy entropy (i.e., randomness) fluctuated significantly, signifying that the agents were still learning. After this initial exploration, the agents tried reward hacking by choosing to take a longer time to traverse the intersection. This is marked by an increase in the episode length (from $\sim$600 to $\sim$700 steps) between 700k and 750k steps. We anticipate this to be an effect of the last reward term, which continuously rewarded the agents inversely proportional to their distance from the goal. However, this phase was quickly overcome, since the probability of collision or lane boundary violations increased and the resulting reward was comparatively insignificant. By now, the policy entropy was starting to settle but was still fluctuating a little. Towards the end of 1M steps, the policy converged at a stable cumulative reward ($\sim$8) and episode length ($\sim$600 steps), while settling at a policy entropy of $\sim$1.2. For LDR and HDR, the KPIs followed a similar trend but with increasing fluctuations. Finally, for FT, we observed an initial performance degradation (owing to the large and sudden domain gap), which eventually recovered and performed well ($>$60\% success rate).

From a computing perspective, we analyzed the effect of parallelizing the intersection-traversal environment from 1 replica (4 agents) up to 25 replicas (100 agents). We observed that the throughput decreased linearly up to 347.8 ms as the replicas increased and that the computational workload was highest for handling physics, followed by MARL, rendering, and least (negligible) for the API/GUI calls (refer Fig. \ref{fig2b}). As shown in Fig. \ref{fig2c}, reduction in training time (up to 76.3\%) was quite non-linear, with a saturating point approaching after 15-20 parallel environments. This resulted in a boost in MARL sample rate from 78.4 Hz for a single replica to 330.7 Hz for 25 parallel environment replicas (refer Fig. \ref{fig2d}).

\subsubsection{Deployment and Sim2Real Transfer}
\label{Sub-Sub-Section: Deployment I}

\begin{figure}[ht]
\includegraphics[width=\linewidth]{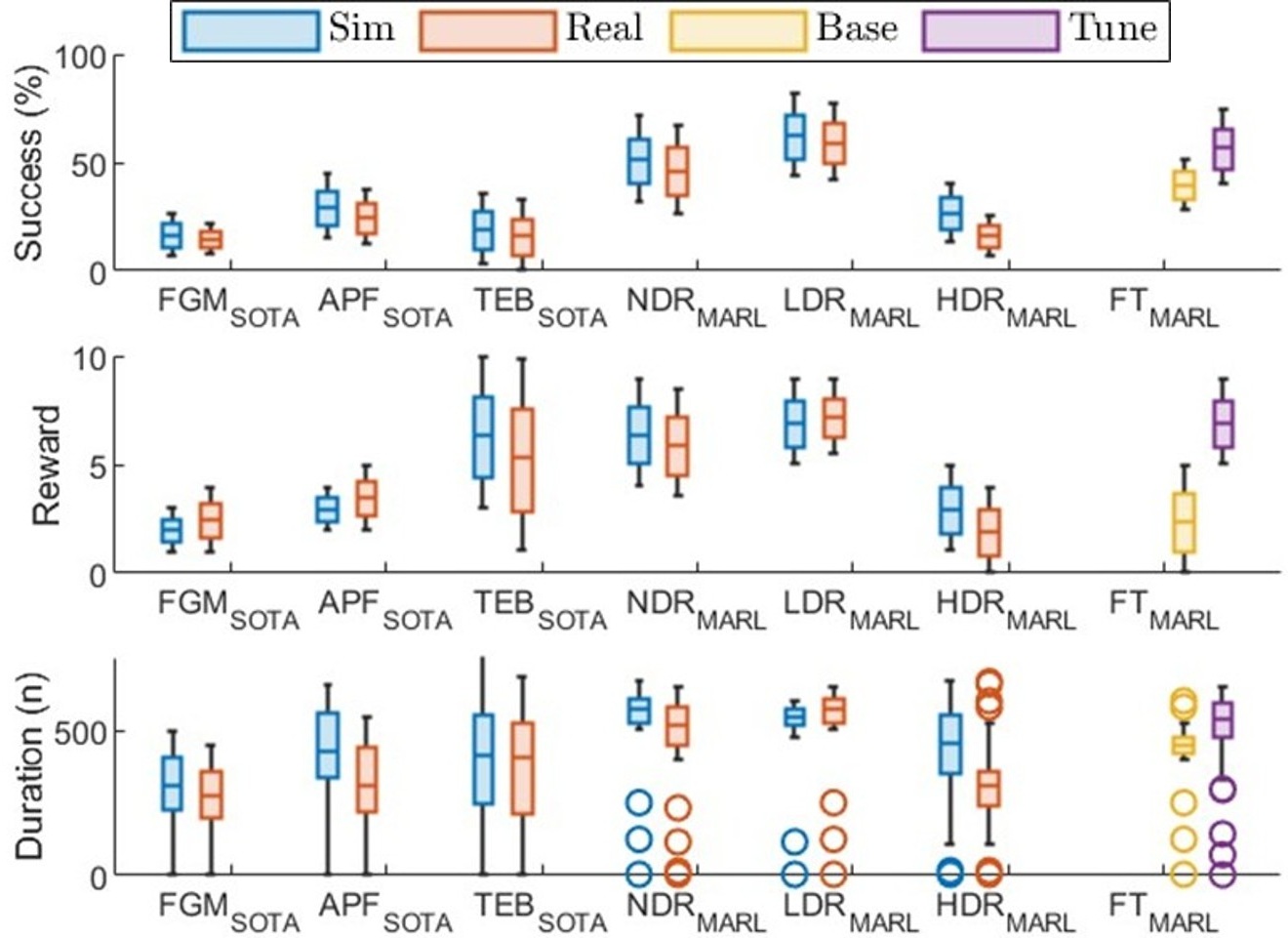}
\caption{Deployment and benchmarking of intersection traversal policies with 4 cooperative agents (A1-A4).}
\label{fig5}
\end{figure}

The trained policies were first deployed and verified in simulation, where we observed interesting emergent behaviors among the agents. The agents strategically slowed down or steered away from each other to avoid collision. Next, we quantitatively analyzed the policies trained with different grades of domain randomization (i.e., NDR, LDR, and HDR) and benchmarked them against FGM, APF, TEB, and FT. The design of experiments followed 16 simulation runs and 16 real-world deployments, where the performance was assessed across 3 KPIs, viz. success rate, cumulative reward, and episode duration aggregated across all the agents (since this was a cooperative scenario) as depicted in Fig. \ref{fig5}. For real-world deployments, our experiments cycled across all the agents, such that each agent was physically deployed in the loop with the simulated environment comprising its virtual peers (refer Section \ref{Sub-Section: Mixed-Reality Digital Twinning} for implementation details).

It was observed that FGM was least successful ($<$30\%), also reflected across the reward ($<$5 points) as well as duration ($<$500 steps) metrics. APF followed with $\sim$40\% success rate, $<$5 reward points, and $<$600 steps. TEB averaged with $\sim$30\% success rate $<$8 reward points, and $<$600 steps, but its performance varied greatly since it returned infeasible solution at times. NDR performed better with mean success rates above 45\%, which allowed it to cash in over 6 reward points on average. Here, the episode duration was between 400 and 650 steps in most cases, with outliers depicting early collisions. LDR was the most consistent with success rates reaching as high as 80\%, rewards as high as 9 points, and episode durations ranging between 500 and 600 steps. HDR performed poorer than other MARL configurations, where it could only achieve up to 40\% success rate. Finally, the performance of base LDR degraded when deployed on turf mat ($<$50\% success), which was improved upon fine-tuning ($\sim$75\% success). It is worth mentioning that the closeness between sim and real metrics estimates the sim2real gap, which was least for LDR (4.12\%), followed by NDR (9.38\%), TEB (13.32\%), FGM (16.01\%), APF (19.41\%), and HDR (33.85\%).

\subsection{Competitive Multi-Agent Scenario}
\label{Sub-Section: Competitive MARL Results}

\begin{figure*}[t]
     \centering
     \begin{subfigure}[b]{0.16\linewidth}
         \centering
         \includegraphics[width=\linewidth]{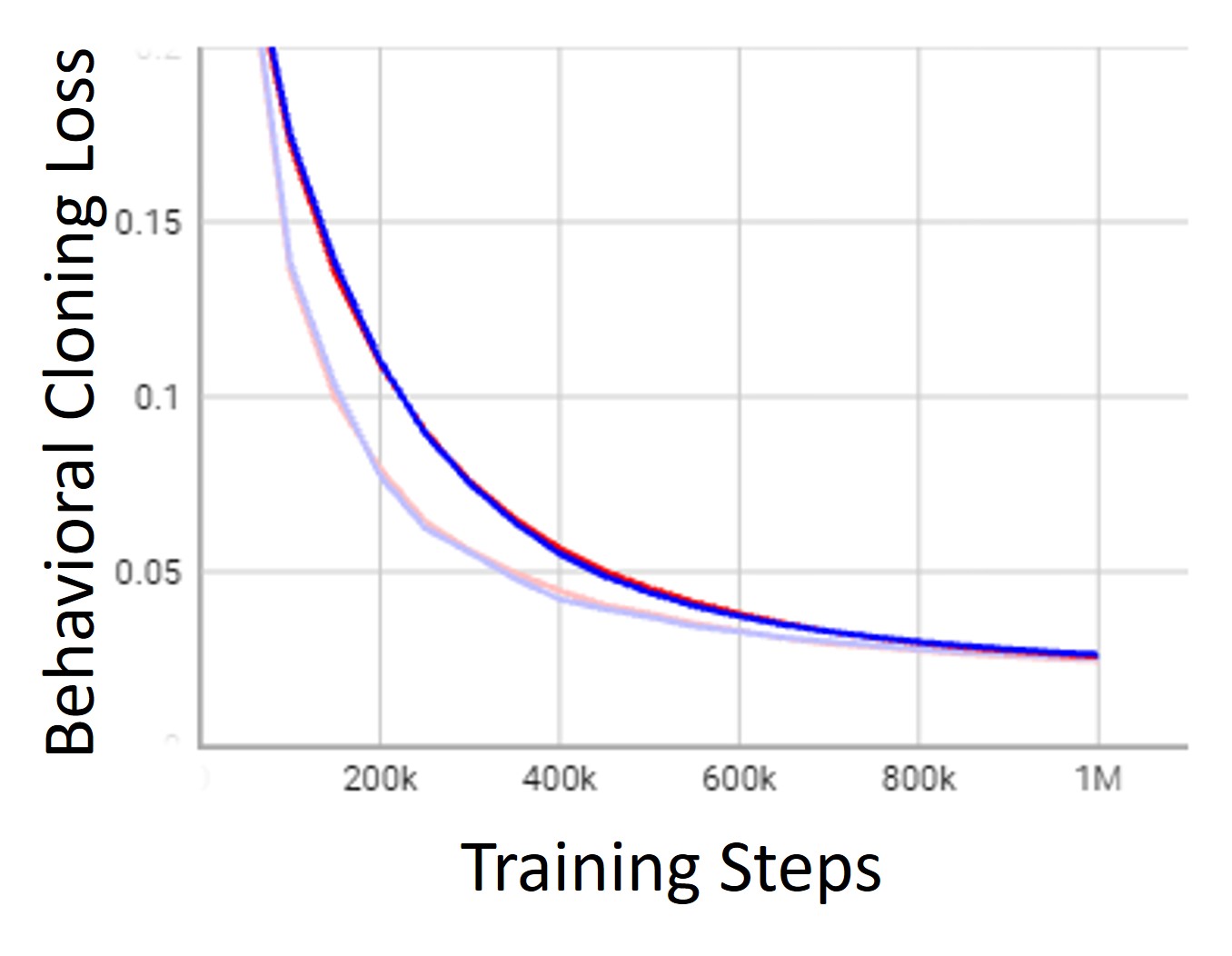}
         \caption{}
         \label{fig6a}
     \end{subfigure}
     \hfill
     \begin{subfigure}[b]{0.16\linewidth}
         \centering
         \includegraphics[width=\linewidth]{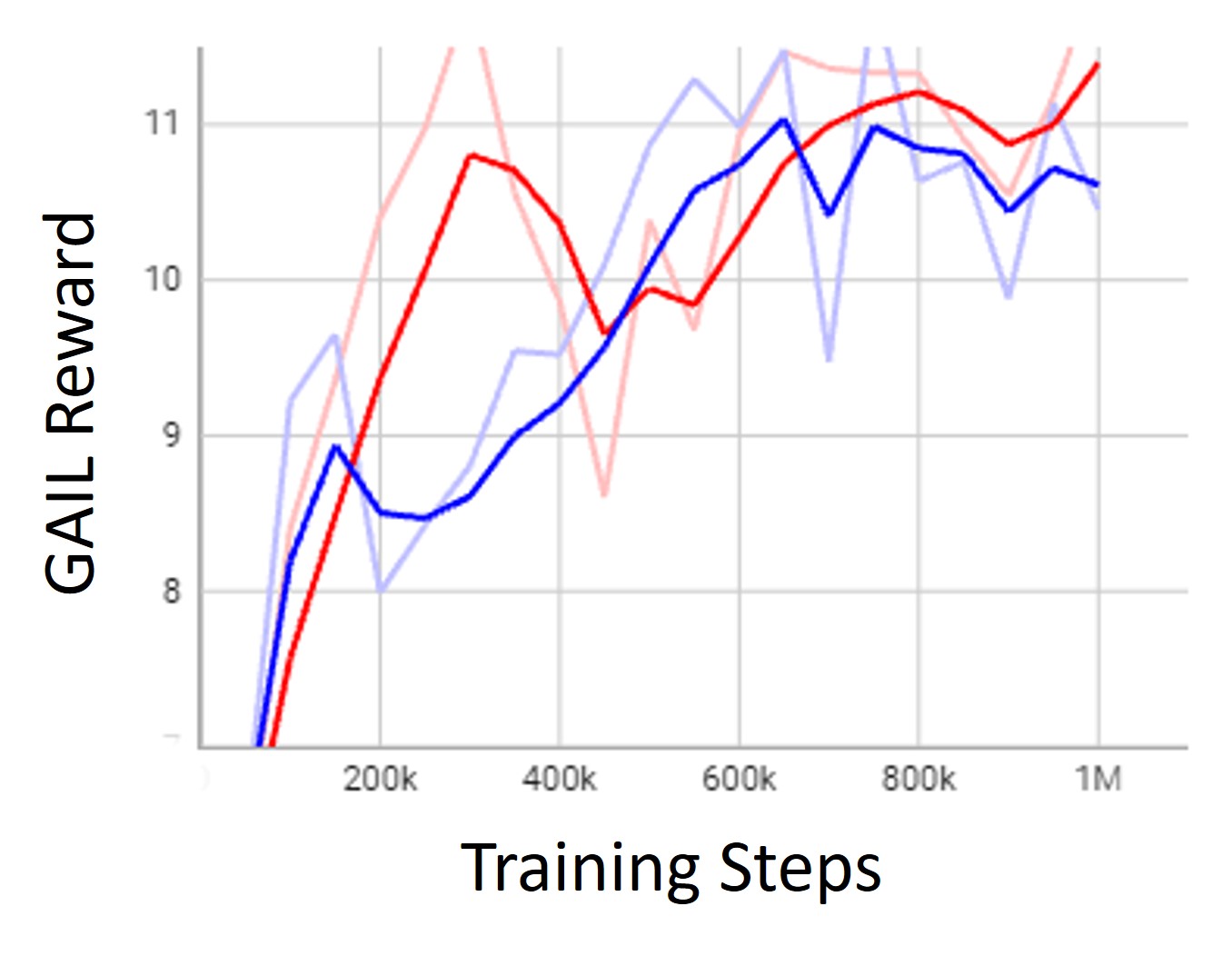}
         \caption{}
         \label{fig6b}
     \end{subfigure}
     \hfill
     \begin{subfigure}[b]{0.16\linewidth}
         \centering
         \includegraphics[width=\linewidth]{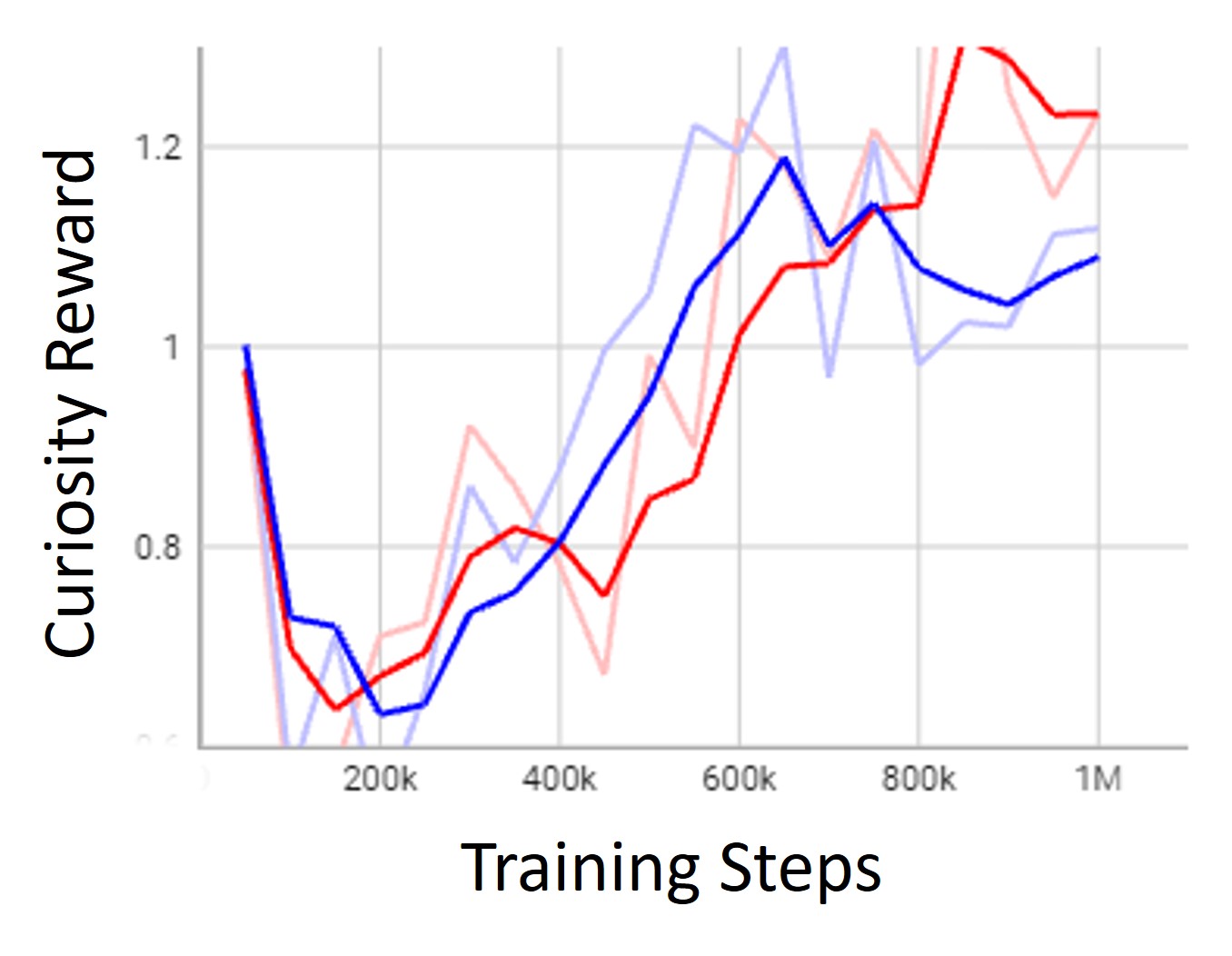}
         \caption{}
         \label{fig6c}
     \end{subfigure}
     \hfill
     \begin{subfigure}[b]{0.16\linewidth}
         \centering
         \includegraphics[width=\linewidth]{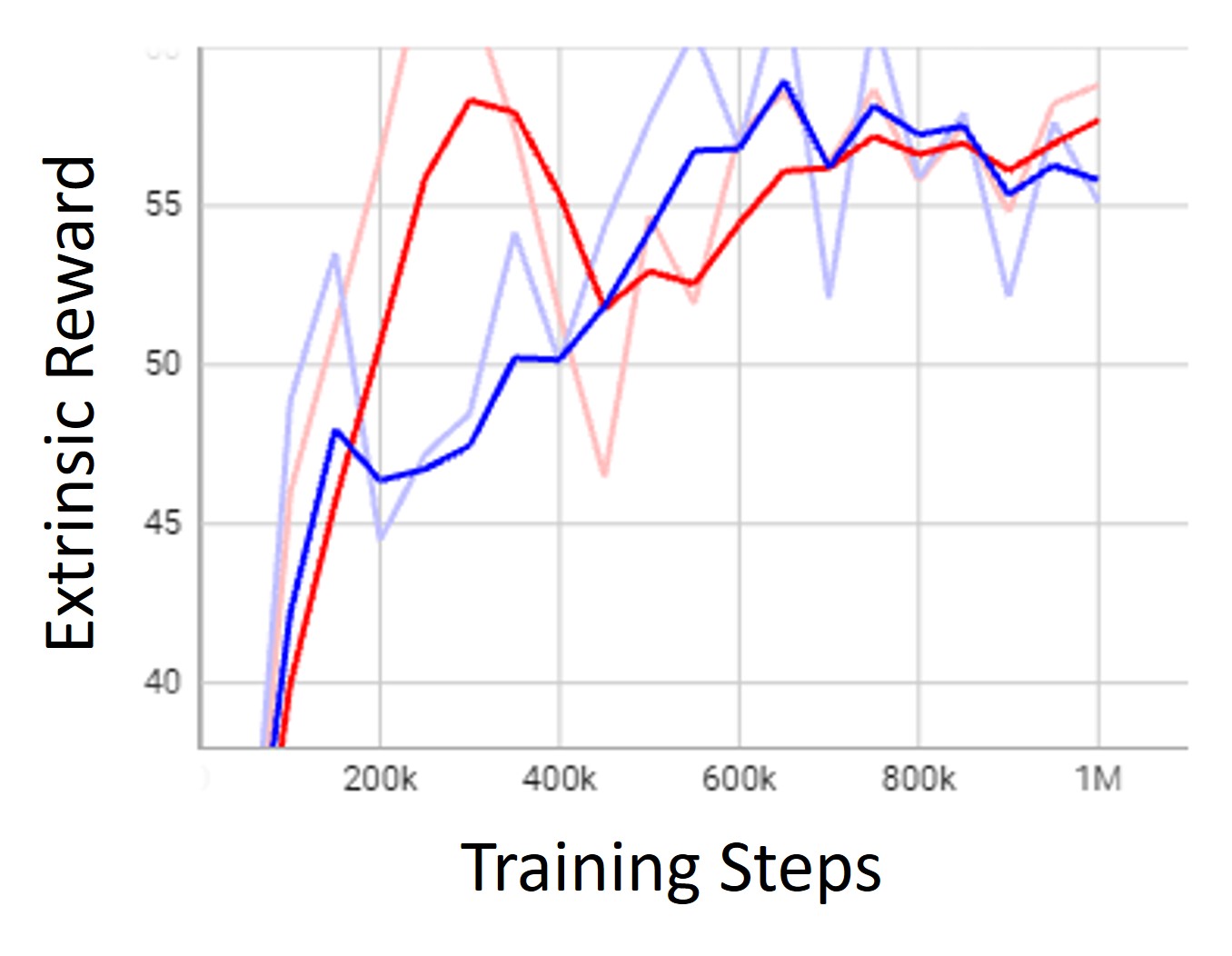}
         \caption{}
         \label{fig6d}
     \end{subfigure}
     \hfill
     \begin{subfigure}[b]{0.16\linewidth}
         \centering
         \includegraphics[width=\linewidth]{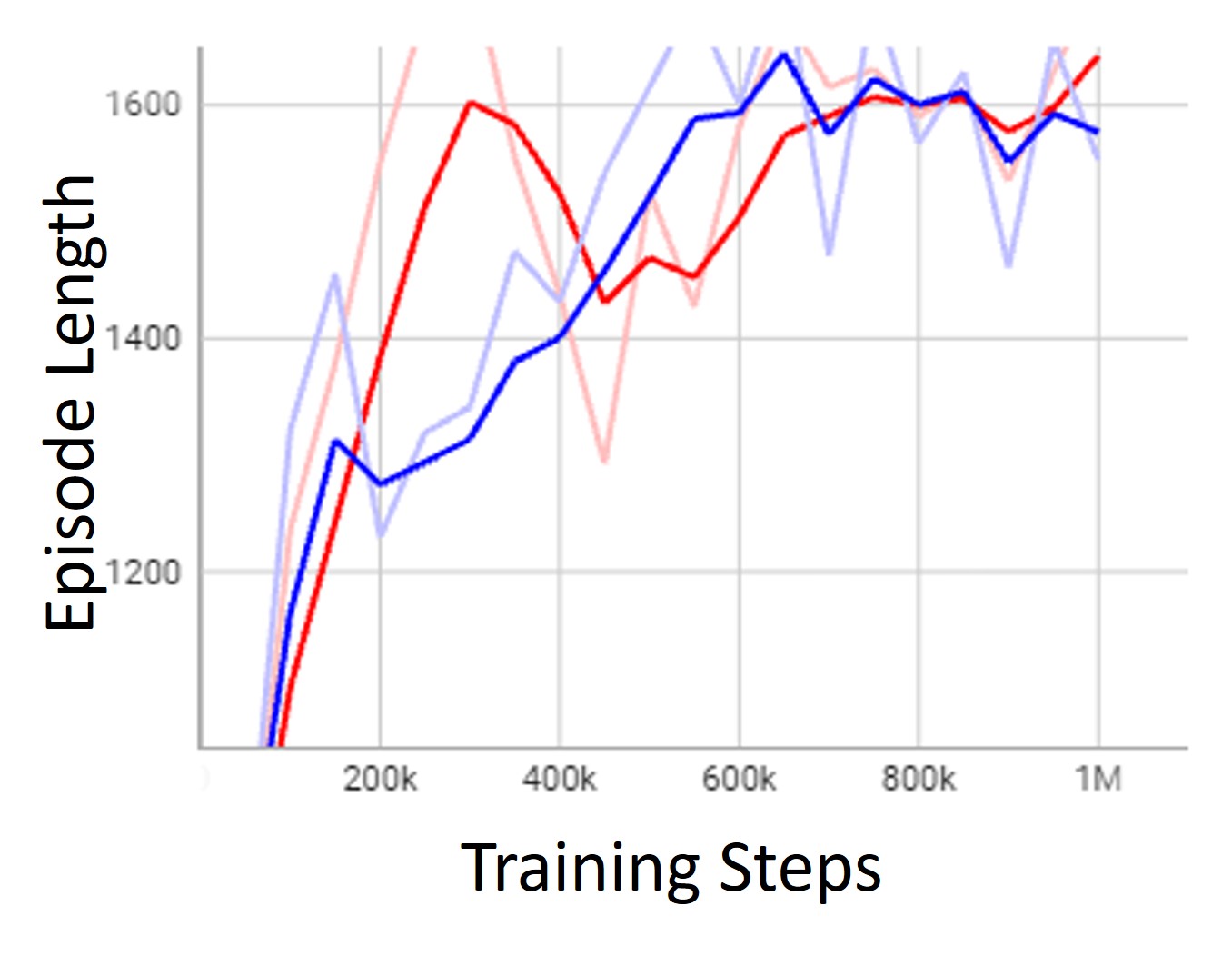}
         \caption{}
         \label{fig6e}
     \end{subfigure}
     \hfill
     \begin{subfigure}[b]{0.16\linewidth}
         \centering
         \includegraphics[width=\linewidth]{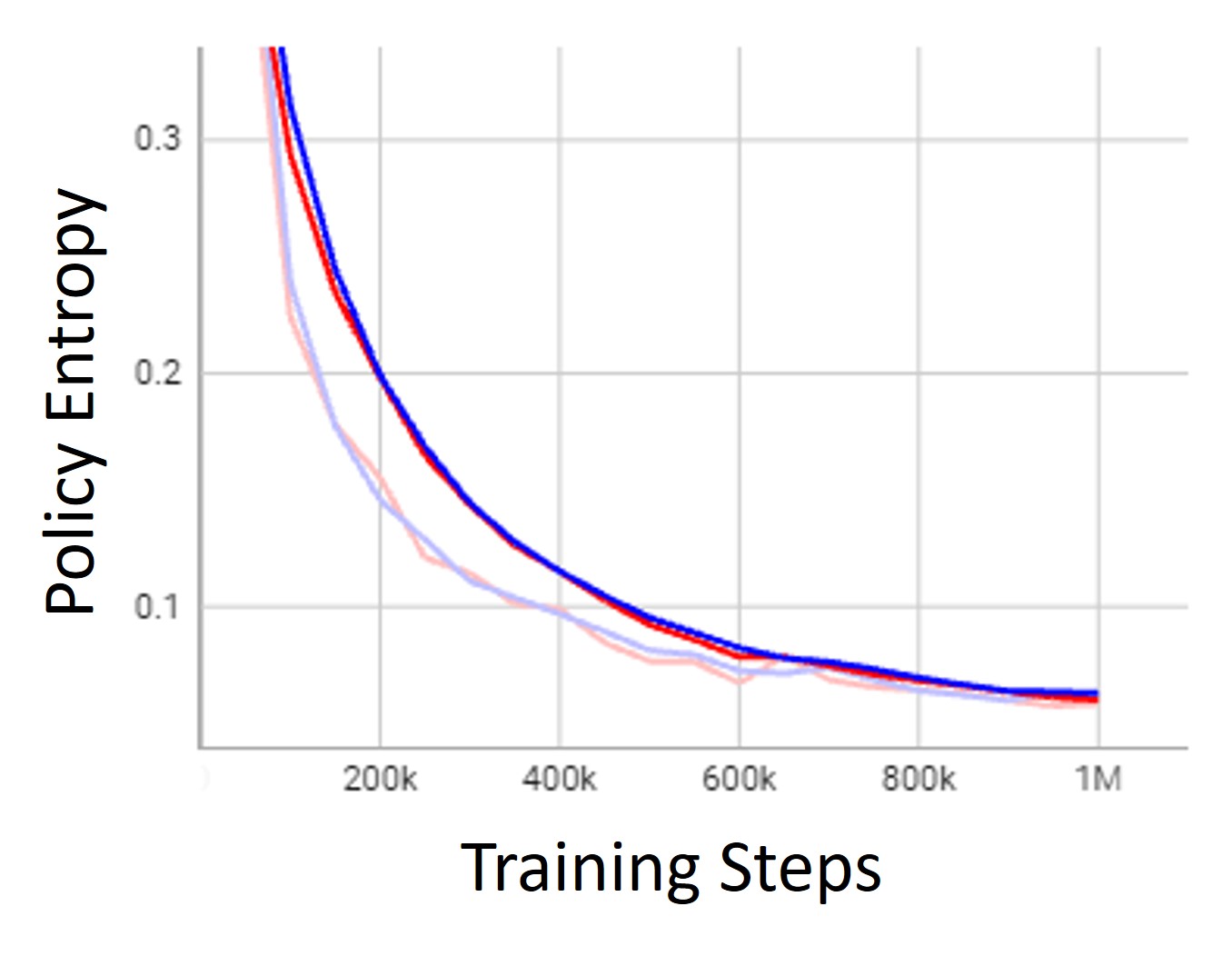}
         \caption{}
         \label{fig6f}
     \end{subfigure}
     \caption{Competitive MARL training: (a) BC loss, (b) GAIL reward, (c) curiosity reward, (d) extrinsic reward, (e) episode length, and (f) policy entropy w.r.t. training steps.}
    \label{fig6}
\end{figure*}

\subsubsection{Training and Simulation Parallelization}
\label{Sub-Sub-Section: Training II}

Fig. \ref{fig6} depicts the KPIs used to analyze the competitive MARL training without any domain randomization (i.e., NDR). It was observed that the agents initially (until $\sim$200k steps) tried aggressive maneuvers, which mostly resulted in collisions. This is marked by the low extrinsic rewards ($<$50) and episode lengths ($<$1400 steps) in this phase. This can also be attributed to the higher BC loss ($>$0.1) as well as lower curiosity ($<$0.8) and GAIL ($<$9) rewards, indicating that the agents had not even started imitating the demonstrations correctly. However, the pre-recorded demonstrations soon (between $\sim$200k and $\sim$800k steps) guided the agents toward completing multiple laps around the race track. This is marked by an exponential reduction in the BC loss (from $\sim$0.1 to $\sim$0.025) and a progressive increase in the extrinsic (from $\sim$50 to $\sim$57), curiosity (from $\sim$0.8 to $\sim$1.1), and GAIL (from $\sim$9 to $\sim$11) rewards as well as the episode length (from $\sim$1400 to $\sim$1600 steps). It was interestingly observed that the red agent (Agent 1) dominated the blue one (Agent 2) till about 500k steps, after which the latter learned the \textit{``competitive spirit''} and bridged the performance gap. Towards the end of 1M steps, both the policies converged at stable reward values and episode length, while gradually reducing the policy entropy from $>$0.3 to $<$0.05. Here, the non-zero offset in BC loss indicates that the agents did not over-fit the demonstrations; rather, they explored the state space quite well to maximize the extrinsic reward by adopting aggressive \textit{``racing''} behaviors. The KPIs followed a similar trend for LDR and HDR, but with higher fluctuations (especially in policy entropy), owing to the randomized parameters.

\begin{figure}[t]
\includegraphics[width=\linewidth]{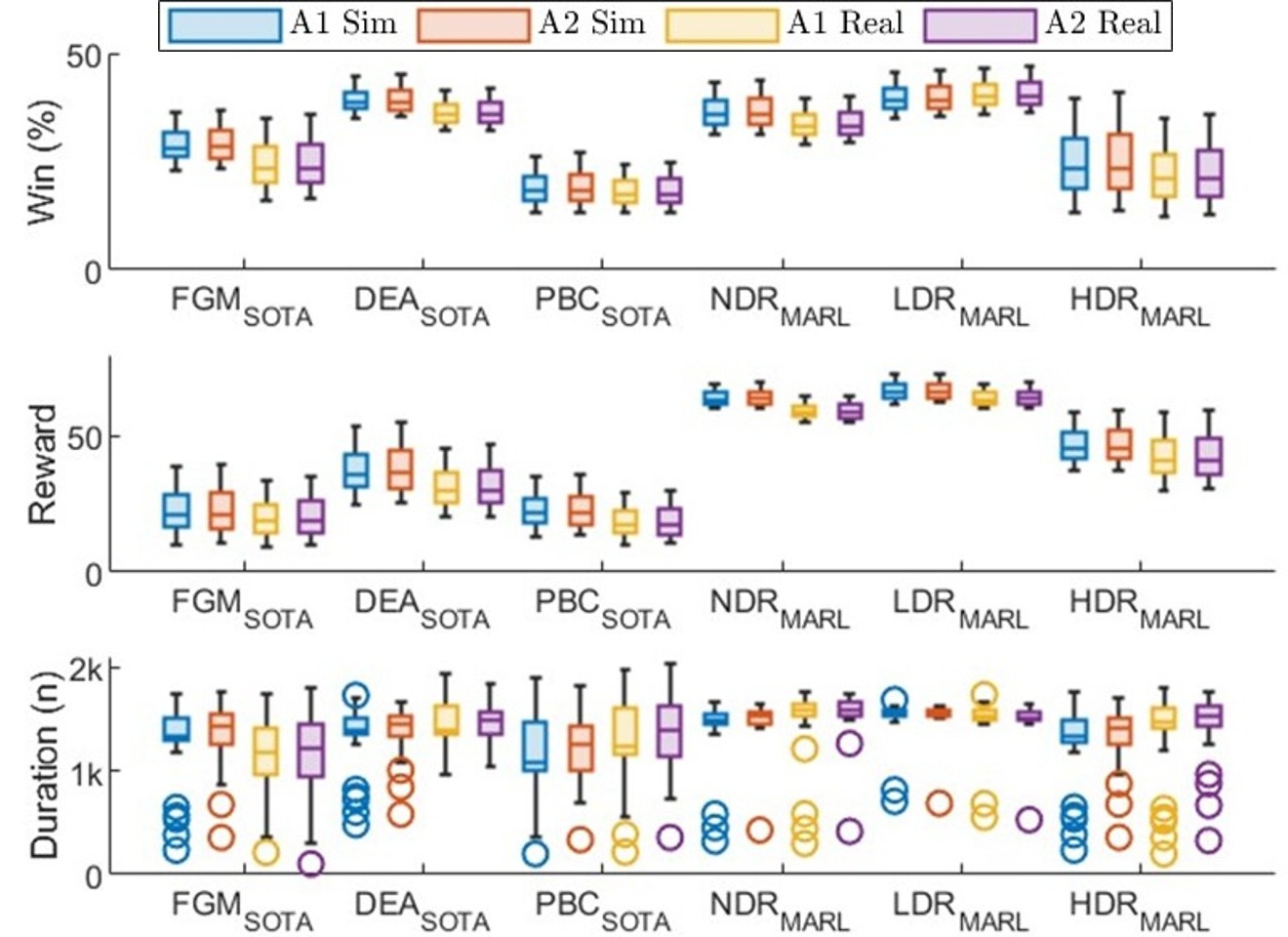}
\caption{Deployment and benchmarking of autonomous racing policies with 2 adversarial agents (A1 and A2).}
\label{fig7}
\end{figure}

From a computing perspective, we analyzed the effect of parallelizing 2-agent adversarial racing family up to 10 such families (20 agents) training in parallel, within the same environment. We observed that the throughput decreased quite non-linearly up to 274.2 ms with a dropping rate of change as the replicas increased (indicating higher computational efficiency compared to environment parallelization scheme) and that the computational workload was highest for handling physics, followed by MARL, rendering, and least for the API/GUI calls (refer Fig. \ref{fig2f}). As observed from Fig. \ref{fig2g}, reduction in training time (up to 49\%) was less dramatic in this case, with a saturating point approaching beyond 10$\times$2 parallel agents. This resulted in a boost in MARL sample rate from 65.9 Hz for a single replica to 120.3 Hz for 10 parallel agent replicas (refer Fig. \ref{fig2h}).

\subsubsection{Deployment and Sim2Real Transfer}
\label{Sub-Sub-Section: Deployment II}

The trained policies were first deployed and verified in simulation, where we observed interesting adversarial behaviors (e.g., blocking, baiting, overtaking, etc.). These behaviors conveyed that the agents were explicitly competing while implicitly coordinating to avoid collisions. Next, we quantitatively analyzed the policies trained with different grades of domain randomization (i.e., NDR, LDR, and HDR) and benchmarked them against FGM, DEA, and PBC. The design of experiments followed 16 simulation runs and 16 real-world deployments, where the performance was assessed across 3 KPIs, viz. win rate, cumulative reward, and episode duration separately for each agent (since this was a competitive scenario) as depicted in Fig. \ref{fig7}. For real-world deployments, our experiments cycled across all the agents, such that each agent was physically deployed in the loop with the simulated environment comprising its virtual peers (refer Section \ref{Sub-Section: Mixed-Reality Digital Twinning} for implementation details).

It was observed that both agents had the lowest average winning rate ($<$25\%) with PBC and HDR, although they secured the least mean reward ($<$25 points) with FGM. This highlighted the difference between losing fairly and losing due to collision, which was also corroborated by the outliers in the duration metric. NDR provided higher ($\sim$30-45\%) winning consistency for either agent, but could not surpass that of LDR ($\sim$35-47\%), which was similar to DEA. The same was reflected by the reward metric, wherein LDR cashed in slightly more reward ($\sim$60-75 points) than NDR ($\sim$55-70 points). Both performed equally well on the duration metric ($\sim$1400-1700 steps), however, LDR was slightly more consistent with lower variance. DEA could collect most reward amongst non-MARL baselines ($\sim$35-45 points) and also performed more consistently. Finally, the sim2real gap was least for LDR (2.88\%), followed by NDR (6.88\%), HDR (8.98\%), DEA (10.82\%), PBC (11.43\%) and FGM (13.48\%).

%%%%%%%%%%%%%%%%%%%%%%%%%%%%%%%%%%%%%%%%%%%%%%%%%%%%%%%%%%%%%%%%%%%%%%%%%%%%%%%%

\section{Conclusion}
\label{Section: Conclusion}

This work identified two pain points in training and deploying MARL systems, and attempted to address them by proposing a scalable and parallelizable digital twin framework. Two representative case studies were formulated to support the claims: a 4-agent collaborative intersection traversal problem and a 2-agent adversarial head-to-head racing problem. The two problems were deliberately formulated with distinct observation spaces and reward functions, but more importantly, also the learning architecture (vanilla MARL vs. demonstration-guided MARL). We analyzed the training metrics in each case and also noted the non-linear effect of agent/environment parallelization on the training time, with a hardware/software-specific point of diminishing return. Finally, we presented a mixed-reality sim2real transfer of the policies (for training/testing) using a single physical vehicle, which was immersed within the proposed digital twin framework to interact with its virtual peers in a virtual environment.

Future avenues of research include analyzing the effect of different communication frameworks and protocols on digital twinning, formulation of physics-guided MARL problems, and scaling the deployments in terms of the number and size of the agents.

%%%%%%%%%%%%%%%%%%%%%%%%%%%%%%%%%%%%%%%%%%%%%%%%%%%%%%%%%%%%%%%%%%%%%%%%%%%%%%%%

%\addtolength{\textheight}{-12cm}  % This command serves to balance the column lengths

%%%%%%%%%%%%%%%%%%%%%%%%%%%%%%%%%%%%%%%%%%%%%%%%%%%%%%%%%%%%%%%%%%%%%%%%%%%%%%%%

\balance
\bibliographystyle{IEEEtran}
\bibliography{references}

\end{document}